\documentclass[10pt,twocolumn,letterpaper]{article}

\usepackage[pagenumbers]{cvpr} 

\usepackage{graphicx}
\usepackage{amsmath}
\usepackage{amssymb}
\usepackage{booktabs}

\usepackage[pagebackref,breaklinks,colorlinks]{hyperref}

\usepackage[capitalize]{cleveref}
\crefname{section}{Sec.}{Secs.}
\Crefname{section}{Section}{Sections}
\Crefname{table}{Table}{Tables}
\crefname{table}{Tab.}{Tabs.}

\def\cvprPaperID{11203} 
\def\confName{CVPR}
\def\confYear{2022}

\begin{document}

\title{Augment to Detect Anomalies with Continuous Labelling}

\author{Vahid Reza Khazaie\\
Western University\\
London, ON, Canada\\
{\tt\small vkhazaie@uwo.ca}
\and
Anthony Wong\\
Western University\\
London, ON, Canada\\
{\tt\small awong655@uwo.ca}

\and
Yalda Mohsenzadeh\\
Western University\\
London, ON, Canada\\
{\tt\small ymohsenz@uwo.ca}

}
\maketitle

\begin{abstract}
Anomaly detection is to recognize samples that differ in some respect from the training observations. These samples which do not conform to the distribution of normal data are called outliers or anomalies. In real-world anomaly detection problems, the outliers are absent, not well defined, or have a very limited number of instances. Recent state-of-the-art deep learning-based anomaly detection methods suffer from high computational cost, complexity, unstable training procedures, and non-trivial implementation, making them difficult to deploy in real-world applications. To combat this problem, we leverage a simple learning procedure that trains a lightweight convolutional neural network, reaching state-of-the-art performance in anomaly detection. In this paper, we propose to solve anomaly detection as a supervised regression problem. We label normal and anomalous data using two separable distributions of continuous values. To compensate for the unavailability of anomalous samples during training time, we utilize straightforward image augmentation techniques to create a distinct set of samples as anomalies. The distribution of the augmented set is similar but slightly deviated from the normal data, whereas real anomalies are expected to have an even further distribution. Therefore, training a regressor on these augmented samples will result in more separable distributions of labels for normal and real anomalous data points. Anomaly detection experiments on image and video datasets show the superiority of the proposed method over the state-of-the-art approaches.

\end{abstract}

\section{Introduction}
\label{sec:intro}

Anomaly detection is the problem of identifying abnormal samples among a group of normal data. This is a deviation from many machine learning problems because the set of abnormal data is either poorly sampled or unavailable during training. Recently, anomaly detection draws huge attention and provides many applications in the field of computer vision like marker discovery in biomedical data \cite{schlegl2017unsupervised} and crime detection in surveillance videos \cite{luo2017revisit}. Tackling these problems involves modelling the distribution of normal visual samples in a way that anomalies are identified at test time.

Deep neural networks have become a popular choice to reach state-of-the-art performance in anomaly detection. Despite their good performance, these models suffer from high computational cost, complexity, and training instability, making them difficult to use in practice. To overcome these limitations, we propose training a relatively shallow CNN with continuous labelling and anomaly creation, yielding state-of-the-art performance on anomaly detection with significantly fewer parameters and less training time. Specifically, we approach anomaly detection as a supervised regression problem, where the model's objective is to map normal and created anomalous data to highly separable distributions.

Due to the unavailability of anomalies, we apply simple data augmentation techniques on normal data to create distinct anomalies. With the new set of anomalous data, we can treat anomaly detection as a supervised learning problem. Since there are now two classes, it is intuitive to treat this as a binary classification problem. However, we show that using regression instead of classification improves anomaly detection performance. Furthermore, we introduce continuous labelling as a favorable means of performance stability.

We evaluated our proposed method, Augment to Detect Anomalies with Continuous Labelling (ADACL), on various benchmark datasets for anomaly detection. ADACL outperforms most state-of-the-art methods using significantly fewer parameters. We also provide a thorough study on loss functions, the choice of labels and the effects of the different augmentations. In this paper, our contributions are the following:

\begin{itemize}
    \item We propose a novel method of anomaly detection which includes a lightweight CNN trained with regression, anomaly creation with augmentations and continuous labelling to improve performance stability. 
    \item Our method is simple yet outperforms most state-of-the-art approaches.
    \item We study the effects of various losses, data augmentations and continuous labelling on anomaly detection performance.
\end{itemize}


\section{Related Works}
\label{sec:related_works}
Several proposed methods such as reconstruction-based approaches take advantage of self-representation learning. They rely on the reconstruction error as a metric to decide whether or not an instance corresponds to the distribution of training examples \cite{xia2015learning, sabokrou2016video}. As such, various types of autoencoders like denoising auotoencoders and context autoencoders \cite{zimmerer2018context} are used for anomaly detection. Most of the deep learning-based models with an autoencoder architecture \cite{sakurada2014anomaly, zhai2016deep, zhou2017anomaly, zong2018deep, chong2017abnormal} also use reconstruction error to detect anomalies. These methods strive to exclusively learn the distribution of normal data in training such that they fail to generalize to anomalies. Even though these methods can be effective in some cases, it has been shown that they generalize well to reconstruct out-of-distribution samples and thus fail to recognize anomalies at testing stage.



Some works used deep convolutional generative adversarial network (DCGAN) \cite{radford2015unsupervised} to learn a manifold of normal images for anomaly detection by mapping from an image space to a random distribution \cite{schlegl2017unsupervised, schlegl2019f}. Sabokrou et al. \cite{sabokrou2018adversarially} proposed a one-class classification framework consisting of a Reconstructor (R) and a Discriminator (D). R serves as a denoising autoencoder, while D operates as the detector. These two networks are trained adversarially in an end-to-end perspective. In an extension to this, Zaheer et al. \cite{zaheer2020old} redefined the adversarial one-class classifier training setup by changing the role of the discriminator to classify between good and bad quality reconstructions and improved the performance even further. In \cite{perera2019ocgan}, Perera et al. leveraged an autoencoder architecture to enforce the normal instances to be distributed uniformly across the latent space. \cite{jewell2021oled} utilized adversarial setup to learn more robust representations by intelligently masking the input.

\cite{salehi2021multiresolution} and \cite{georgescu2021anomaly} have attempted to benefit from deep pre-trained networks by distilling the knowledge where a small student model learns from a large teacher model. In \cite{salehi2021multiresolution}, they utilized a VGG-16 \cite{simonyan2014very} to calculate a multi-level loss from different activations for training the student network to determine the anomaly score. They also incorporate interpretability algorithms in their framework to localize anomalous regions and perform anomaly segmentation. Although knowledge-distillation methods could perform anomaly detection with high performance, they benefit from pre-training on millions of labeled images which is not effective in other modalities of data. Also, in practice, knowledge-distillation methods may not be suitable due to computationally expensive inference. Our proposed method does not leverage pre-trained networks, so we do not compare against these approaches.

Gong et al. proposed a deep autoencoder augmented with a memory module \cite{gong2019memorizing} to encode the input to a latent space with the encoder. The resulting latent vector is used as a query to retrieve the most relevant memory item for reconstruction with the decoder. Also, in \cite{park2020learning}, they introduced a memory module with items that capture prototypical models of the inlier class with a new update system.

Some of the proposed anomaly detection methods that rely on learning the distribution of inliers cannot be applied to real-world applications. Generating anomalies alongside the available normal data build an informative training set for the task of anomaly detection. Employing GANs for generating anomalous data turns the problem of anomaly detection into a binary classification problem. This method can also be used for data augmentation for anomalous data. In \cite{pourreza2021g2d}, they trained a Wasserstein GAN on normal samples and utilized the generator before convergence. In this case, generated irregular data have a controlled deviation from inliers. Although they set a new research direction in anomaly detection, training a network to generate outliers is computationally expensive.

\section{Method}
\label{sec:method}

\subsection{Motivation}
Deep neural networks have shown great performance in solving anomaly detection problems. However, they often have a large number of parameters, making them difficult and expensive to train. Not only that, most deep models suffer from training instability, complexity, difficult implementation, and is intractable to deploy in real-world applications. To overcome these issues, we propose ADACL, where we follow an intuitive and stable training procedure which also exceeds state-of-the-art performance. ADACL is simple to implement and has relatively few parameters, leading to inexpensive training and fast inference time. Therefore, it is more suitable for use in real world scenarios. 

\subsection{Approach}

To improve performance in anomaly detection, it is desirable to produce representations of normal and anomalous samples that have distinct distributions. In our method, we redefine anomaly detection as a supervised regression problem. However, the training data consists mainly of normal samples, which makes supervised learning a cumbersome task. To solve this issue, we leverage straightforward data augmentations to create anomalous samples during training. 

\subsubsection{Regression for Anomaly Detection}
\begin{figure}[htbp]
\begin{center}
   \includegraphics[width=1\linewidth]{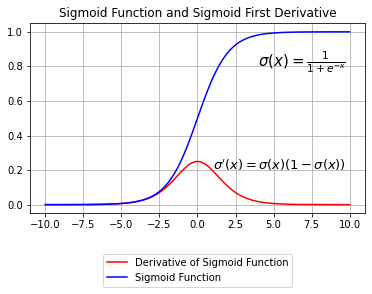}
\end{center}
   \caption{Sigmoid function and the first derivitive of the sigmoid function. Rate of change of sigmoid decreases to 0 as predictions approach its target.}
\label{sigandfirstsig}
\end{figure}

\begin{figure}[htbp]
\begin{center}
   \includegraphics[width=1\linewidth]{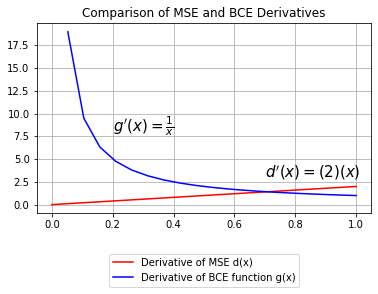}
\end{center}
   \caption{The first derivates of MSE and BCE. Notice that the gradients of MSE become larger than BCE after a certain point.}
\label{bcemsederiv}
\end{figure}

We utilize a lightweight convolutional neural network (CNN) to train on the normal and created anomalies. The CNN acts as a regressor that outputs a continuous value between 0 and 1 to represent normal and anomalous data. Even though this can be considered a binary classification problem, we show that regression offers fast convergence and high performance in anomaly detection. Here we will explain the different configuration options we have for our method. As examples, we explain why we did not use sigmoid in the last layer of the CNN and instead used value clipping as well as why Mean Squared Error (MSE) was chosen over Binary Cross Entropy (BCE) as the loss function. In the case of binary classification, equation \ref{sigmoidgrowth} shows that the rate of change of the sigmoid function is always decreasing as the prediction approaches the ground truth target. Also, as shown in figure \ref{sigandfirstsig}, the value of gradients are nearing zero in the same manner. Due to the saturation of gradients near the target, updating the weights will be less effective, resulting in slow convergence. A similar problem exists in binary classification with Binary Cross Entropy (BCE). Negative log likelihood in BCE also exhibits a decreasing rate of change as the predicted value approaches the ground truth target. Referring to figure \ref{bcemsederiv},  Mean Squared Error (MSE) provides stronger gradients as the prediction approaches the target, which results in faster convergence. We perform anomaly detection experiments on ADACL and find not only that MSE converges faster, but also manages to maintain consistently high performance across multiple training runs. Consequently, we transform anomaly detection into a regression problem to reach the optimal solution faster.


\begin{equation}
\begin{aligned}
h(x) =  \frac{\mathrm{1} }{\mathrm{1} + e^{-x} } 
\end{aligned}
\end{equation}

\begin{equation}
\begin{aligned}
h'(x) = h(x)(1-h(x))
\end{aligned}
\end{equation}

\begin{equation}
\begin{aligned}
for \ x < 0: \qquad h'(x) - h'(x-1) > 0 \\
for \ x > 0: \qquad h'(x-1) - h'(x) > 0\\
\end{aligned}
\label{sigmoidgrowth}
\end{equation}

\begin{figure*}[htbp]
\begin{center}
   \includegraphics[width=1\linewidth]{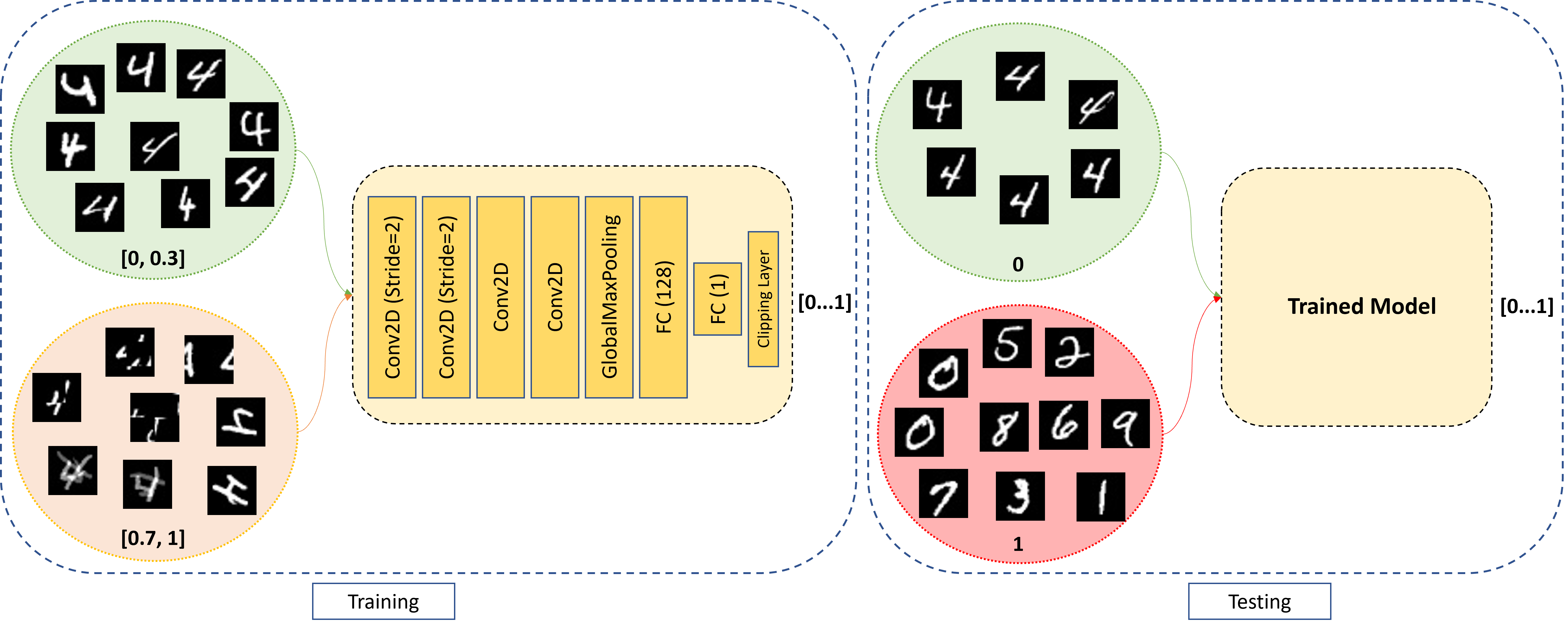}
\end{center}
   \caption{Overview of ADACL architecture. Normal examples and created anomalies are assigned continuous labels and fed into the CNN. This regression model outputs a continuous value between 0 and 1 as an anomaly score. At test time, the model is evaluated with real anomalies.}
\label{modelarch}
\end{figure*}

\subsubsection{Continuous Labelling}
In the anomaly detection problem, we can label normal and anomalous data as 0 and 1, respectively. We call this Discrete Labelling (DL). However, experimental results show that this leads to high variance in anomaly detection performance. Instead, we use Continuous Labelling (CL), where we designate two continuous intervals corresponding to normal and anomalous data, and sample labels from them using a uniform distribution. The intuition behind continuous labelling is that the expected value of MSE over predictions is lower in comparison to using discrete labelling. Let a discrete label $ \in \{0, 1\}$ and a continuous label $ \in \{[0, X_L], [X_H, 1]\}$. $X_L$ is the upper bound of the interval of normal class and $X_H$ is the lower bound of the interval of anomaly class. Because we sample from a uniform distribution, $A = \mathbb{E}([0, X_L]) = \frac{X_L}{2}$ and $B = \mathbb{E}([X_H, 1]) = \frac{X_H + 1}{2}$. A and B are the expected values of prediction for the normal and anomaly classes, respectively. The MSE function takes two numbers as the input to calculate the loss. As shown in equation \ref{expectedmse}, we let the prediction of our model be 0.5 (highest distance to the lower and upper bounds). The following inequalities show that the value of the MSE loss is always lower when using continuous labelling compared to discrete labelling:

\begin{equation}
\begin{aligned}
MSE(0.5, A) < MSE(0.5, 0) \ if \ A > 0 \\
MSE(0.5, B) < MSE(0.5, 1) \ if \ 1 > B
\\
\end{aligned}
\label{expectedmse}
\end{equation}

Therefore:

\begin{equation}
\begin{aligned}
\mathbb{E}(MSE(prediction, CL)) <\\ \mathbb{E}(MSE(prediction, DL))
\end{aligned}
\label{expected}
\end{equation}

According to equation \ref{expected}, if we choose continuous labelling over discrete labelling, the expected value for the loss is lower during training and thus, convergence is slower. Therefore, it should increase training stability. The experimental results in Figures \ref{ValVarianceCIFAR} and \ref{AUCVarianceCIFAR} supports this hypothesis.

\begin{figure}[htbp]
\begin{center}
   \includegraphics[width=1\linewidth]{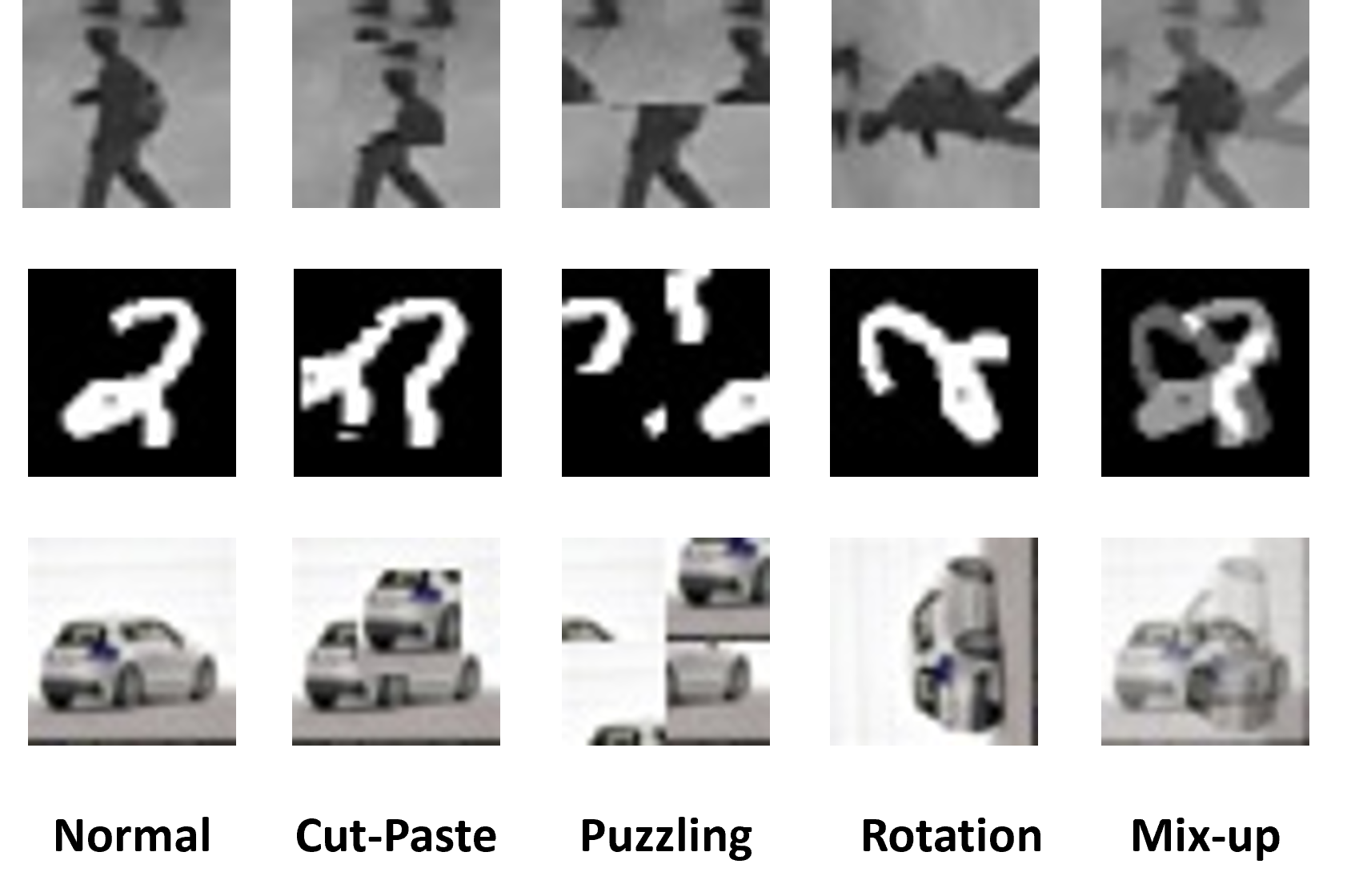}
\end{center}
   \caption{Various Augmentations applied to create anomalies. The first, second, and third rows contain normal and created anomalies with different augmentations of images from UCSD, MNIST and CIFAR-10, respectively. }
\label{augs}
\end{figure}

\subsubsection{Anomaly Creation}
Solving anomaly detection as a supervised regression problem requires a dataset containing both normal and anomalous data. To compensate for the unavailability of anomalies, we utilize data augmentation techniques to create them during training. Examples of these augmentations are shown in figure \ref{augs}. The following are descriptions of our proposed augmentations for ADACL:

\begin{itemize}
    \item \textbf{\textit{Cut-Paste:}} Randomly select patch from image and place it in a random location. 
    \item \textbf{\textit{Puzzling:}} Take quarters of the image and shuffle them.
    \item \textbf{\textit{Rotation:}} Rotate the image 90 degrees one or three times
    \item \textbf{\textit{Mix-up:}} Add a rotated image to the original one. Prior to adding, the rotated and original image are multiplied by respective coefficients. 
\end{itemize}

To assign training labels to normal and created anomalies, we pick two separate continuous intervals from which we uniformly sample. For example, normal and anomalous labels are in the range of [0, 0.3] and [0.7, 1], respectively.

\subsection{Implementation Details}
Our method which is depicted in figure \ref{modelarch}, uses a simple CNN with less than 300k parameters. As a regressor, this model outputs a continuous value which is clipped between 0 and 1. It is trained on the inlier samples and created anomalies which are augmented versions of the normal data. We use different variations of the Adam optimizer in conjunction with a cyclic learning rate. Also, we designate a constant number of epochs for training on each dataset. Then, based on the validation set, we use early stopping techniques to terminate training. This validation set consists of 150 randomly selected samples of normal data and the augmented version of them as anomalies. This random selection maintains consistency by using a random seed.  

\section{Experiments and Results}
\label{sec:exp_res}
\subsection{Anomaly Detection in Images}

The image datasets we choose for anomaly detection in this paper are MNIST \cite{lecun1998mnist}, FMNIST \cite{xiao2017fashion} and CIFAR-10 \cite{krizhevsky2009learning}. These benchmark datasets are standard in anomaly detection literature. In the following, we provide descriptions and protocols defined on each dataset. 
\par
{\bf MNIST:} It is a dataset of handwritten digits that has 60,000 images. Samples in MNIST are grayscale with a resolution of 28 x 28. This is a benchmark dataset in anomaly detection.
\par
{\bf FMNIST:} Fashion MNIST also contains 60,000 28 × 28 grayscale images of fashion accessories but since there is a significant amount of intra-class variation, it is a more challenging dataset compared to MNIST.

{\bf CIFAR-10:} This dataset consists of 10 classes of 32 × 32 RGB images of natural objects. With high intra-class variance, CIFAR-10 is a more challenging benchmark for anomaly detection.

\par
\begin{figure}[htbp]
\begin{center}
   \includegraphics[width=0.7\linewidth]{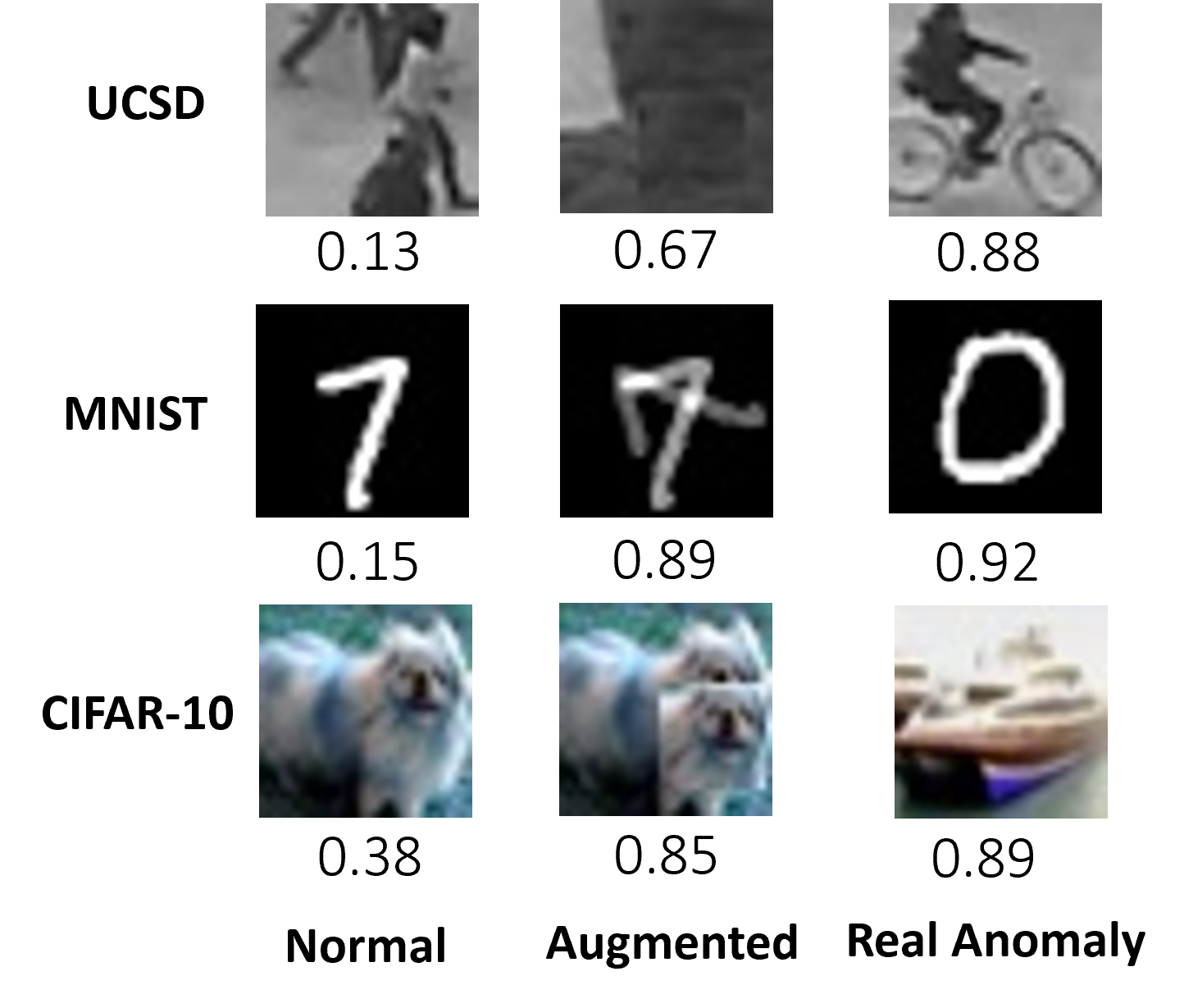}
\end{center}
   \caption{Predictions of the model on normal, augmented and anomalous samples from UCSD, MNIST, and CIFAR-10.}
\label{normalauganomexamples}
\end{figure}
\begin{figure}[htbp]
\begin{center}
   \includegraphics[width=1\linewidth]{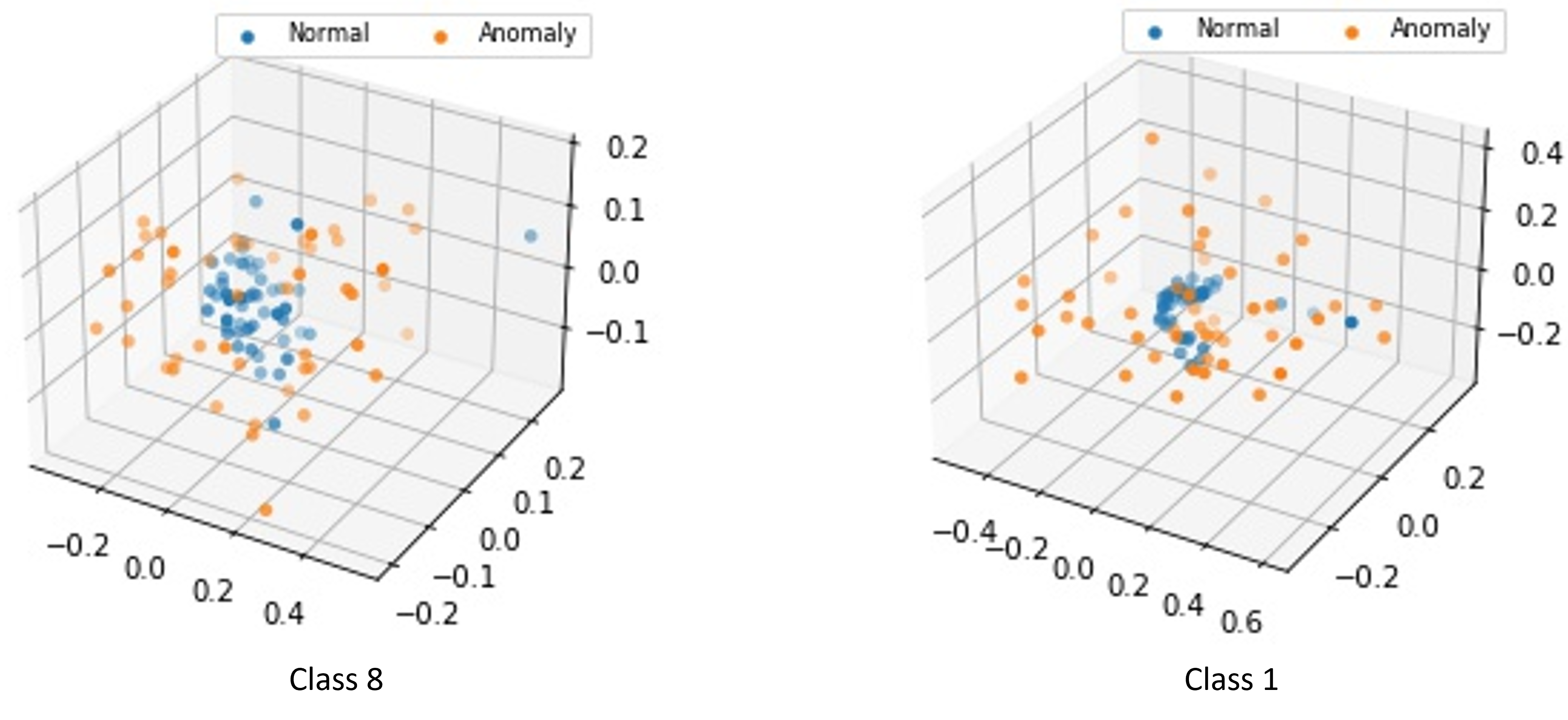}
\end{center}
   \caption{3D visualization of learned representation of class 1 and 8 of the MNIST dataset. As shown, there are separable and distinct distributions of normal and anomalous embeddings.}
\label{learned_repr}
\end{figure}

The protocol we follow for these three datasets is to consider one class as normal data and the rest as anomalies. To measure anomaly detection performance, we calculate Area Under the Curve (AUROC) for each class and report the average of all classes as the final performance. AUROC on these datasets are shown in table \ref{table:mnist,fashionmnist,CIFAR-10}. From our results, ADACL outperforms recent state-of-the-art anomaly detection methods. Moreover, figure \ref{normalauganomexamples} depicts the model's predictions on normal, augmented and anomalous samples over different datasets. Figure \ref{learned_repr} shows the 3D distributions of learned representations of normal and anomalous samples on class 1 and 8 of MNIST. This figure shows the separability of learned representations.

\begin{table*}[htbp]
\centering
\caption{AUROC in \% for anomaly detection on MNIST \cite{lecun1998mnist},  Fashion-MNIST \cite{xiao2017fashion} and CIFAR-10 \cite{krizhevsky2009learning} datasets.}
\label{table:mnist,fashionmnist,CIFAR-10}
\resizebox{\textwidth}{!}{\begin{tabular}{|c|c|c|c|c|c|c|c|c|c|c|c|c|}
\hline
Dataset & Method & 0 & 1 & 2 & 3 & 4 & 5 & 6 & 7 & 8 & 9 & Mean\\

\hline

MNIST
& AnoGAN\cite{schlegl2017unsupervised} & 96.6 & 99.2 & 85.0 & 88.7 & 89.4 & 88.3 & 94.7 & 93.5 & 84.9 & 92.4 & 91.3\\
& DSVDD\cite{ruff2018deep} & 98.0 & 99.7 & 91.7 & 91.9 & 94.9 & 88.5 & 98.3 & 94.6 & 93.9 & 96.5 & 94.8\\
& OCSVM\cite{scholkopf2002learning} & 99.5 & 99.9 & 92.6 & 93.6 & 96.7 & 95.5 & 98.7 & 96.6 & 90.3 & 96.2 & 96.0\\
& CapsNet\textsubscript{PP} \cite{li2020exploring} & 99.8 & 99.0 & 98.4 & 97.6 & 93.5 & 97.0 & 94.2 & 98.7 & 99.3 & 99.0 & 97.7\\
& OCGAN\cite{perera2019ocgan} & 99.8 & 99.9 & 94.2 & 96.3 & 97.5 & 98.0 & 99.1 & 98.1 & 93.9 & 98.1 & 97.5\\
& LSA\cite{abati2019latent} & 99.3 & 99.9 & 95.9 & 96.6 & 95.6 & 96.4 & 99.4 & 98.0 & 95.3 & 98.1 & 97.5\\

& \textbf{Ours (ADACL)} & ${99.37}$ & ${99.30}$ & ${98.58}$ & ${97.36}$ & ${97.57}$ & ${98.43}$ & ${99.56}$ & ${98.09}$ & ${93.46}$ & ${98.38}$ & \textbf{98.01}\\

\hline
Dataset & Method & T-shirt & Trouser & Pullover & Dress & Coat & Sandal & Shirt & Sneaker & Bag & Ankle boot & Mean\\

\hline

Fashion-MNIST
& DAGMM\cite{zong2018deep} & 30.3 & 31.1 & 47.5 & 48.1 & 49.9 & 41.3 & 42.0 & 37.4 & 51.8 & 37.8 & 41.7\\
& DSEBM\cite{zhai2016deep} & 89.1 & 56.0 & 86.1 & 90.3 & 88.4 & 85.9 & 78.2 & 98.1 & 86.5 & 96.7 & 85.5\\
& LSA\cite{abati2019latent} & 91.6 & 98.3 & 87.8 & 92.3 & 89.7 & 90.7 & 84.1 & 97.7 & 91.0 & 98.4 & 92.2\\
& DSVDD\cite{ruff2018deep} & 98.2 & 90.3 & 90.7 & 94.2 & 89.4 & 91.8 & 83.4 & 98.8 & 91.9 & 99.0 & 92.8\\
& OCSVM\cite{scholkopf2002learning} & 91.9 & 99.0 & 89.4 & 94.2 & 90.7 & 91.8 & 83.4 & 98.8 & 90.3 & 98.2 & 92.8\\

& \textbf{Ours (ADACL)} & ${94.42}$ & ${99.46}$ & ${89.82}$ & ${91.05}$ & ${92.68}$ & ${90.40}$ & ${80.43}$ & ${97.88}$ & ${97.14}$ & ${98.88}$ & \textbf{93.22}\\

\hline
Dataset & Method & Plane & Car & Bird & Cat & Deer & Dog & Frog & Horse & Ship & Truck & Mean\\

\hline

CIFAR-10
& OCSVM\cite{scholkopf2002learning} & 63.0 & 44.0 & 64.9 & 48.7 & 73.5 & 50.0 & 72.5 & 53.3 & 64.9 & 50.8 & 58.56\\
& CapsNet\textsubscript{PP}\cite{li2020exploring} & 62.2 & 45.5 & 67.1 & 67.5 & 68.3 & 63.5 & 72.7 & 67.3 & 71.0 & 46.6 & 61.2\\
& AnoGAN\cite{schlegl2017unsupervised} & 67.1 & 54.7 & 52.9 & 54.5 & 65.1 & 60.3 & 58.5 & 62.5 & 75.8 & 66.5 & 61.79\\
& DSVDD\cite{ruff2018deep} & 61.7 & 65.9 & 50.8 & 59.1 & 60.9 & 65.7 & 67.7 & 67.3 & 75.9 & 73.1 & 64.81\\
& LSA\cite{abati2019latent} & 73.5 & 58.0 & 69.0 & 54.2 & 76.1 & 54.6 & 75.1 & 53.5 & 71.7 & 54.8 & 64.1\\
& OCGAN\cite{perera2019ocgan} & 75.7 & 53.1 & 64.0 & 62.0 & 72.3 & 62.0 & 72.3 & 57.5 & 82.0 & 55.4 & 65.66\\
& CAVGA-D\textsubscript{u}\cite{venkataramanan2020attention} & 65.3 & 78.4 & 76.1 & 74.7 & 77.5 & 55.2 & 81.3 & 74.5 & 80.1 & 74.1 & 73.7\\
& DROCC\cite{goyal2020drocc} & 81.66 & 76.74 & 66.66 & 67.13 & 73.62 & 74.43 & 74.43 & 71.39 & 80.02 & 76.21 & 74.23\\

& \textbf{Ours (ADACL)} & ${73.89}$ & ${83.87}$ & ${67.47}$ & ${70.66}$ & ${69.51}$ & ${77.91}$ & ${72.66}$ & ${83.04}$ & ${87.64}$ & ${81.35}$ & \textbf{76.80}\\

\hline

\end{tabular}}
\end{table*}

Table \ref{table:mnist,CIFAR-10} compares our results with those of two other methods that use the knowledge distillation framework. Methods such as these rely on pre-trained networks which have been trained on millions of labelled images. To learn from these pre-trained or teacher networks, these methods have been trained over many epochs. These methods are computationally expensive and require a long time for inference, which prevents their use in real-world scenarios. Our method takes less time and computation to train even though the results are slightly lower as shown in the table.

\begin{table*}[htbp]
\centering
\caption{Comparison of AUROC in \% for anomaly detection on MNIST \cite{lecun1998mnist} and CIFAR-10 \cite{krizhevsky2009learning} datasets with knowledge distilation methods.}
\label{table:mnist,CIFAR-10}
\resizebox{\textwidth}{!}{\begin{tabular}{|c|c|c|c|c|c|c|c|c|c|c|c|c|c|}
\hline
Dataset & Method & 0 & 1 & 2 & 3 & 4 & 5 & 6 & 7 & 8 & 9 & Mean & Epoch\\

\hline

MNIST

& U-Std\cite{bergmann2020uninformed} & 99.9 & 99.9 & 99 & 99.3 & 99.2 & 99.3 & 99.7 & 99.5 & 98.6 & 99.1 & 99.35 & -\\

& Multiresolution KDAD \cite{salehi2021multiresolution} & 99.82 & 99.82 & 97.79 & 98.75 & 98.43 & 98.16 & 99.43 & 98.38 & 98.41 & 98.1 & 98.71 & 50\\

& Ours (ADACL) & ${99.37}$ & ${99.30}$ & ${98.58}$ & ${97.36}$ & ${97.57}$ & ${98.43}$ & ${99.56}$ & ${98.09}$ & ${93.46}$ & ${98.38}$ & 98.01 & 10\\

\hline

Dataset & Method & Plane & Car & Bird & Cat & Deer & Dog & Frog & Horse & Ship & Truck & Mean & Epoch\\

\hline

CIFAR-10

& U-Std\cite{bergmann2020uninformed} & 78.9 & 84.9 & 73.4 & 74.8 & 85.1 & 79.3 & 89.2 & 83 & 86.2 & 84.8 & 81.96 & -\\

& Multiresolution KDAD \cite{salehi2021multiresolution} & 90.53 & 90.35 & 79.66 & 77.02 & 86.71 & 91.4 & 88.98 & 86.78 & 91.45 & 88.91 & 87.18 & 200\\

& Ours (ADACL) & ${73.89}$ & ${83.87}$ & ${67.47}$ & ${70.66}$ & ${69.51}$ & ${77.91}$ & ${72.66}$ & ${83.04}$ & ${87.64}$ & ${81.35}$ & 76.80 & 15\\

\hline

\end{tabular}}
\end{table*}

\begin{table}
\centering
\caption{Frame-level AUCROC and EER comparison \% on UCSD dataset with state-of-the-art methods.}

 \begin{tabular}{|l|l|c|}
 \hline
Method & AUCROC (\%) & EER (\%) \\ [0.5ex] 
 \hline
TSC \cite{luo2017revisit_novelty}                       & 92.2 & -                  \\
FRCN action \cite{hinami2017joint_novelty}               & 92.2                  & -                  \\
AbnormalGAN \cite{ravanbakhsh2017abnormal_novelty}               & 93.5                   & 13               \\
MemAE \cite{gong2019memorizing} & 94.1                   & -                \\
GrowingGas \cite{sun2017online}                & 94.1   & -                \\
FFP \cite{liu2018future_novelty}     & 95.4  & -                  \\
ConvAE+UNet \cite{Nguyen_2019_ICCV}               & 96.2  & -                  \\
STAN \cite{lee2018stan}        & 96.5   &   -    \\
Object-centric \cite{ionescu2019object} &  97.8   & -                       \\ 
Ravanbakhsh \cite{ravanbakhsh2019training} & - & 14       \\
ALOCC \cite{sabokrou2018adversarially} & - & 13       \\
Deep-cascade \cite{sabokrou2017deep_novelty} & - & 9       \\
 Old is gold \cite{zaheer2020old} & 98.1 & 7       \\

 \textbf{Ours (ADACL)}  &  \textbf{98.4} &  \textbf{7} \\
 
 \hline
\end{tabular}

\label{ucsd_exp}
\end{table}

\subsection{Anomaly Detection in Videos}
To evaluate our method on video anomaly detection, we selected the UCSD dataset \cite{xiao2017fashion}. This dataset contains multiple outdoor scenes with mobile objects such as pedestrians, cars, wheelchairs, skateboards and bicycles. Frames with only pedestrians are considered as the normal class, while frames containing other objects are anomalies. This dataset contains two subsets named Ped1 and Ped2. Ped1 includes 34 training video samples and 36 testing video samples and Ped2 contains 2,550 frames in 16 training videos and 2,010 frames in 12 test videos with a resolution of 240 × 360 pixels.
\par
We follow a patch-based protocol to evaluate on this dataset where each frame is divided into 30 x 30 sections. For training, we include only patches that include pedestrians. However, the model was evaluated on patches that contain pedestrians or other objects. Following \cite{zaheer2020old}, to report the performance on this dataset, frame-level Area Under the Curve (AUROC) and Equal Error Rate (EER) are calculated in table \ref{ucsd_exp}. Results show that ADACL surpasses state-of-the-art methods for video anomaly detection on UCSD.  

\subsection{Analysis of the Method}
In this section, we provide experimental results that support the intuition behind our method.

\subsubsection{BCE vs. MSE}
\begin{figure}[htbp]
\begin{center}
   \includegraphics[width=1\linewidth]{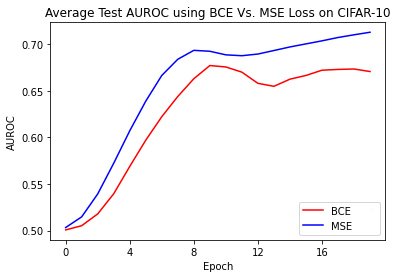}
\end{center}
   \caption{ Average test AUROC taken over 10 training runs on CIFAR-10. MSE achieves higher AUROC in a shorter number of epochs. }
\label{AUCmsebce}
\end{figure}

Based on our experimental results, we use MSE over BCE because it converges faster but still has good performance over multiple training runs. The figure \ref{AUCmsebce} shows that our experimental results are aligned with this hypothesis.

\subsubsection{Continuous vs. Discrete Labelling}
\begin{figure}[htbp]
\begin{center}
   \includegraphics[width=1\linewidth]{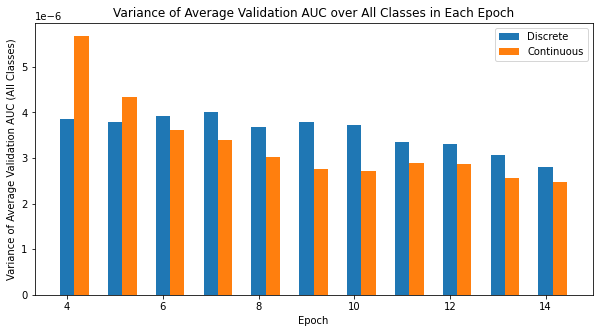}
\end{center}
   \caption{Variance of Average Validation AUROC over all classes in each epoch. The averages are taken over 10 training runs. It can be observed that continuous labelling consistently produces lower variance in higher epochs. Thus, a stopping criteria based on the validation set yields more stable test AUROC when continuous labels are used.}
\label{ValVarianceCIFAR}
\end{figure}

\begin{figure}[htbp]
\begin{center}
   \includegraphics[width=1\linewidth]{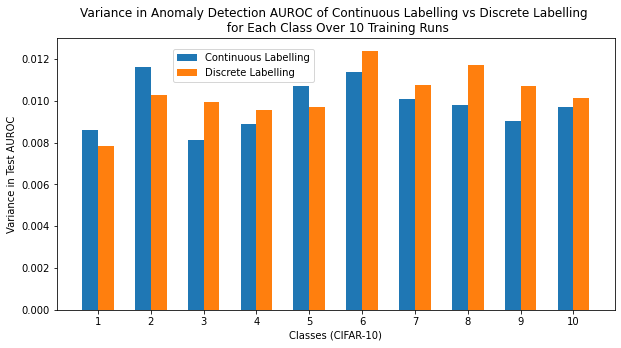}
\end{center}
   \caption{Variance in Anomaly Detection AUROC when using continuous labelling versus discrete labelling for all class taken over 10 training runs of CIFAR-10. More often than not, using the same stopping criteria with continuous labels produces lower variance in test AUROC. }
\label{AUCVarianceCIFAR}
\end{figure}

Referring back to the method section, we define an early stopping criteria based on validation AUROC. In this experiment, we study how labelling affects anomaly detection performance. We keep the entire training procedure the same and only modify the labelling scheme. Results show that discrete labels cause higher variance in the validation and test AUROC. With higher instability, it is increasingly difficult to create an accurate stopping criteria. As shown in figure \ref{ValVarianceCIFAR}, continuous labelling yields lower variance in validation AUROC. Knowing this, figure \ref{AUCVarianceCIFAR} shows that with continuous labelling, a stopping criteria over a validation set with lower AUROC variance is mostly able to produce more consistent test AUROC values per class. Therefore, continuous labelling is the better choice when using a stopping criteria for anomaly detection.

\begin{table}
\centering
\caption{Experiments on different intervals}
 \begin{tabular}{|c|c|c|}
 \hline
Interval & Mean AUROC (\%) & Variance \\ [0.5ex] 
 \hline
$[0, 0.1] - [0.9, 1]$  & $97.14$ & $1.50 \times 10^-5$                  \\
$[0, 0.2] - [0.8, 1]$ & $96.93 $    & $8.83 \times 10^-5 $                \\
$[0, 0.3] - [0.7, 1]$ & $97.42 $  & $1.29 \times 10^-5 $      \\
 
 \hline
\end{tabular}

\label{interval_exp}
\end{table}

\subsubsection{Effects of Augmentations}
\begin{figure}[htbp]
\begin{center}
   \includegraphics[width=1\linewidth]{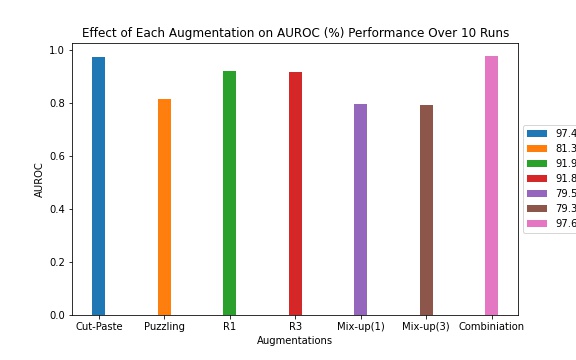}
\end{center}
   \caption{The effects of the augmentations like Cut-paste, puzzling, one and three times rotation and various mix-ups on AUROC. The most effective augmentation was Cut-paste,
   however using all augmentations in the process of creating outliers yields the highest performance in anomaly detection.}
\label{effect_augs}
\end{figure}

Variations in anomaly detection performance occur when using data augmentation to create anomalies. In this experiment, we study the effect of each augmentation used to create anomalies during training. As seen in figure \ref{effect_augs}, the best performing solo augmentation is Cut-paste. But, the best anomaly detection results are achieved by using all augmentations. 

\subsubsection{Interval Selection}
In previous experiments, we show that continuous labelling improves anomaly detection by enabling better early stopping. This is achieved through lower variance in validation AUROC. To further examine the implications of label selection, We analyze different sized intervals to see their effects on anomaly detection performance. As shown in table \ref{interval_exp}, the choice of interval has low impact on AUROC. 

\section{Conclusion}
Deep neural networks can achieve state-of-the-art performance when applied to anomaly detection tasks. However, most of them suffer from expensive computations, high complexity, training instability, and difficult implementation. In this paper, we alleviate these issues by proposing a simple and effective methodology for anomaly detection. We convert the problem into a supervised regression task by creating anomalies using data augmentations and training a lightweight convolutional neural network over continuous labels. In further experiments, we analyze the effects of MSE Loss versus BCE Loss, continuous labelling, interval size, and various augmentations. Results on several image and video anomaly detection benchmarks show our superiority over cutting-edge methods. 

{\small
\bibliographystyle{ieee_fullname}
\bibliography{egbib}
}

\end{document}